# Unsupervised Extraction of Phenotypes from Cancer Clinical Notes for Association Studies

Stefan G. Stark[1,3,4,5], Stephanie L. Hyland[1-5], Melanie F. Pradier[1,7,8], Kjong-Van Lehmann[1,3,4,5], Andreas Wicki[9,10], Fernando Perez-Cruz[7,11,+], Julia E. Vogt[3,5,12,+], Gunnar Rätsch[1-6,+]

**Affiliations:**
1 Computational Biology Program, Memorial Sloan Kettering Cancer Center, New York, U.S.A.
2 Tri-Institutional Ph.D. Program in Computational Biology and Medicine, Weill Cornell Medicine, New York, U.S.A.
3 Department of Computer Science, ETH Zürich, Zürich, Switzerland
4 Medical Informatics Group, University Hospital Zürich, Zürich, Switzerland
5 Swiss Institute for Bioinformatics, Zurich, Switzerland
6 Department of Biology, ETH Zürich, Zürich, Switzerland
7 Department of Signal Processing and Information Theory, University Carlos III in Madrid, Leganés, Spain
8 School of Engineering and Applied Sciences, Harvard University, Cambridge, MA, U.S.A.
9 Department of Biomedicine, University of Basel, Basel, Switzerland
10 Tumorzentrum, University Hospital Basel, Basel, Switzerland
11 Swiss Data Science Center, ETH Zürich and EPFL Lausanne, Switzerland
12 Department of Mathematics and Computer Science, University of Basel, Basel, Switzerland
[+] These authors jointly directed this work. Contact: gunnar.ratsch@ratschlab.org, jvogt@inf.ethz.ch, fernando.perezcruz@sdsc.ethz.ch

## Abstract

The recent adoption of Electronic Health Records (EHRs) by healthcare providers has introduced an important source of data that provides detailed and highly specific insights into patient phenotypes over large cohorts. These datasets, in combination with machine learning and statistical approaches, generate new opportunities for research and clinical care. However, many methods require the patient representations to be in structured formats, while the information in the EHR is often locked in unstructured text designed for human readability.
In this work, we develop the methodology to automatically extract clinical features from clinical narratives from large EHR corpora without the need for prior knowledge. We consider medical terms and sentences appearing in clinical narratives as atomic information units. We propose an efficient clustering strategy suitable for the analysis of large text corpora and utilize the clusters to represent information about the patient compactly. Additionally, we define the sentences on ontologic and natural language vocabularies to automatically detect pertinent combinations of concepts present in the corpus, even when an ontology is not available.
To demonstrate the utility of our approach, we perform an association study of clinical features with somatic mutation profiles from 4,007 cancer patients and their tumors. We apply the proposed algorithm to a dataset consisting of ≈65 thousand documents with a total of ≈3.2 million sentences. After correcting for cancer type and other confounding factors, we identify a total of 340 significant statistical associations between the presence of somatic mutations and clinical features. We annotated these associations according to their novelty and we report several known associations. We also propose 37 plausible, testable hypothesis for associations where the underlying biological mechanism does not appear to be known. These results illustrate that the automated discovery of clinical features is possible and the joint analysis of clinical and genetic datasets can generate appealing new hypotheses that may give rise to new research connecting molecular with clinical features.

# Introduction

Electronic Health Records (EHR) are a vast and valuable data source able to describe patient states to a high degree of granularity. Recently, EHR adoption at healthcare centers has become commonplace, aggregating individual patient experiences into large datasets. Such datasets, when combined with advancements in machine learning methods, introduce new and exciting opportunities for healthcare research and applications. Models and algorithms are now able to scale to these datasets, and by leveraging large cohort sizes, find interactions previously hidden. In particular, clinical decision support systems promise improved patient monitoring and care, personalized medicine introduces the ability to tailor therapies to the individual, and genetic association studies provide insight into the effects of dangerous genetic mutations[1–3]. These methods all benefit from patient representations of increasing detail; however, these representations are required to be in machine-readable, computable formats, and capturing fine-grained and relevant features into such a format is challenging.

To extract information contained in unstructured clinical narratives, researchers have applied techniques from natural language processing[4–7]. Algorithms now exist to parse text documents, identify medical concepts, and map them to codes representing these concepts. This process of detecting concepts in free-text is known as named entity recognition (NER), while matching them to codes is called normalization[3,8,9]. Codes may be organized in an ontology or a language system that enumerates natural language phrases and often further defines their relationships. To standardize biomedical vocabularies, the Unified Medical Language System (UMLS) repository defines an ontology of medical concepts, comprised of codes called Concept Unique Identifiers (CUI)[8].

Phenotype-genotype association studies have been proposed to take advantage of the increasing availability of genetic information and EHR-derived phenotypes. The definition of phenotype and genotype can be quite broad, but for this work, we consider the phenotype as the presence or absence of a clinical observation. Different from most association studies involving germline variants[10], we consider the genotype as the presence or absence of a somatic mutation[11,12]. Generally, association studies correlate patient phenotypes, potentially extracted from the EHR, to patient genotypes, as a tool to study mutational effects in genetic diseases not necessarily limited to cancer[1,13–15]. Early versions of these association studies in cancer helped discover genetic risk factors that contributed to increased risks of cancers such as breast[16], and lung[17], leading to therapies targeting these genetic factors. Studies of most common cancer types have now been performed, detecting over 450 germline variants associated with an increased cancer risk[15]. Advancements in genome sequencing technologies, such as highly scalable next-generation sequencing (NGS) methods, has made the genome, and in particular the cancer genome, much more accessible[18]. Healthcare providers routinely sequence patients, generating an abundant source of mutational data, and potentially pair them to the patient EHR. In particular, Zehir et al[19] described the efforts at a large cancer center to routinely profile tumor and healthy tissue to detect somatic mutations of the tumors.

EHR adoption coupled with natural language processing techniques such as NER has created the opportunity to extract and study more granular phenotypes, moving beyond high-level phenotypes such as cancer type. These studies often require expert-assisted analysis to verify results, due to inherent biases in cohort selection, and confounding effects arising from the genotypes and more detailed phenotypes. Results are often used to drive research by generating hypotheses, to be tested with controlled experiments[3]. Such studies have primarily relied on structured data, for example, Denny et. al.[20] performed an association study between ICD codes and four mutations in the gene FOXE1, finding significant associations with thyroiditis, nodular and multinodular goiters, and thyrotoxicosis. Hebbring et. al.[21] explored text-based phenotypes as an alternative to ICD codes and extract n-grams from a corpus, filter out rare occurrences and perform associations against five SNP mutations with effects known a priori. Through an association study, Denny et. al.[22] identified a novel gene that participates in a cardiac rhythm condition. The presence of this heart condition was detected by an algorithm that extracted concepts from EHR text and combined them with heart vital signs and other structured data. Moreover, Chan et al.[23] explored the use of topic models to summarize cancer clinical notes and associate them with somatic mutations.

## Results

**Development of a new method to identify clinical phenotypes**

In this work, we propose a method to capture interesting and relevant features from a large text corpus in an unsupervised manner. The proposed method first clusters sentences from clinical narratives contained in a large EHR corpus and represents a patient by a set of these sentence clusters. A sentence in a clinical narrative is considered a single observation by a healthcare professional. By clustering them our method creates a structured form of data that is able to represent patients as a set of these clinical observations, also it automatically detects phrases and combinations of clinical concepts that occur in the text (**Figure 1 top**).

To demonstrate the utility of our approach, we use obtained clusters as phenotypes in an association study. Somatic mutations[19] were matched to the patient clinical narratives. We conduct association studies between three phenotype classes, one that represents patients by the occurrence CUI tokens detected in their text and two that represent patients by the occurrences of sentences clusters, using the two sets of previously described clusters. Finally, these associations are analyzed and expert-annotated according to their novelty, reporting several known associations and proposing plausible and testable hypotheses for a subset of currently unexplainable associations (**Figure 1 bottom**).

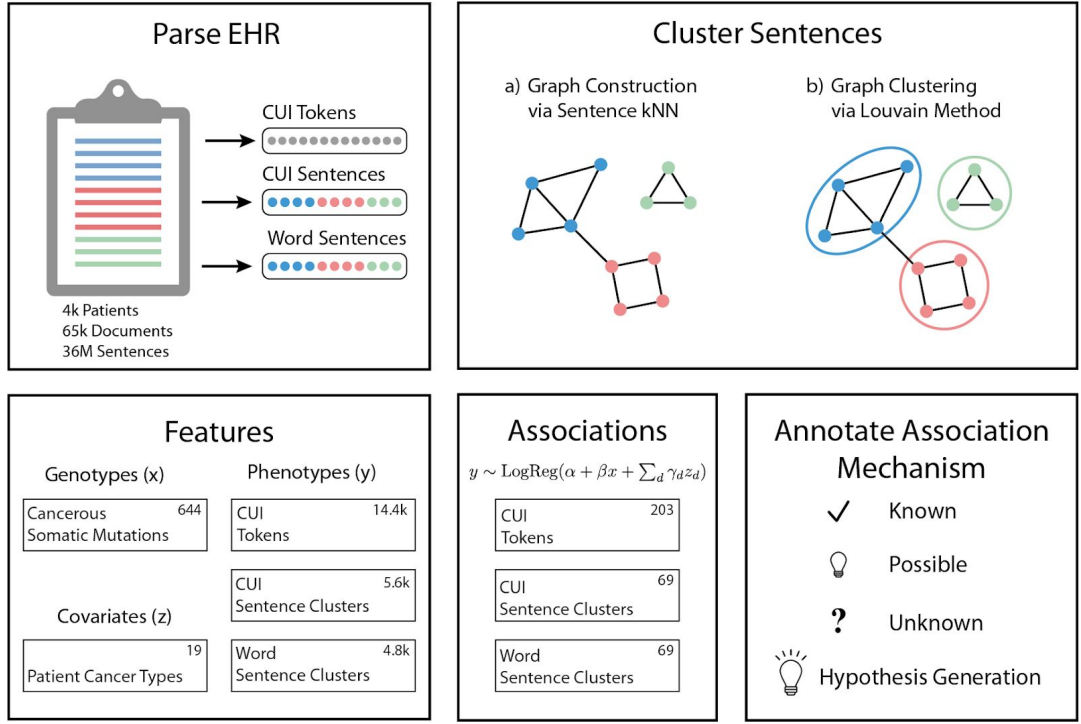

**Figure 1:** We consider a cohort of 4,007 cancer patients. Sentences from about 65k EHR documents are parsed using two parsers. The first uses the MetaMap parser to detect UMLS codes, referred to as CUI tokens, contained in the sentences; the second employs a simple natural language parser to create "Word sentences". The two sets of sentences are clustered by constructing and clustering a kNN graph. The construction of this graph proceeds with an efficient kNN search procedure. We extract 3 phenotypes from the EHR corpus, representing patients as sets of CUI tokens, CUI sentence clusters, and Word sentence clusters. Somatic mutations were available for each patient[19] and we performed an association study with the three extracted phenotypes. These identified associations were annotated according to the novelty of their mechanism and were used to formulate testable hypotheses about the connection between somatic mutations and clinical phenotypes.

## Extraction of clinical phenotypes from cancer clinical narratives

The dataset used for this work contains information from 4,007 cancer patients, with clinical narratives from the EHR matched to somatic mutations taken from tumor samples. In total there are 65,147 documents, with a median of 7 per patient. Across all documents are 3,203,698 sentences, with a median of 44 sentences per document. Using the word-parser there we identified 1,009,062 unique sentences with a median sentence length of 7 words. Using the CUI-parser, we identified 792,023 unique sentences with a median sentence length of 5 tokens.

We applied the proposed methodology (Figure 1) to the dataset described above and obtained 14.4k CUI tokens, 5.6k clusters of CUI sentence clusters and 4.8k word sentence clusters. Information about the clusters is provided as supplementary material (Supplementary Table 2). For each patient, we then computed high-dimensional representations where each dimension corresponds to a previously defined cluster. The dimension *j* for a patient *i* indicates the occurrence of sentences associated to cluster *j* in clinical notes for patient *i* (see Methods for more details).

When we use the so-derived high-dimensional representation of the patient's EHR records to visualize the patients, we find that the representations exhibit a complex structure that is reflective of the cancer types of the patients' tumors (**Figure 2**). We observe that patients with the same or similar cancer types usually appear in the same region in a t-SNE visualization[24], indicating that the representation is at least carrying the information about the cancer type. The information about the cancer type (as provided by the pathologist) was not used to determine the clustering and the patient representation. Figure 2 (right) also shows examples of cluster prototype sentences that frequently appear in patients with that cancer type.

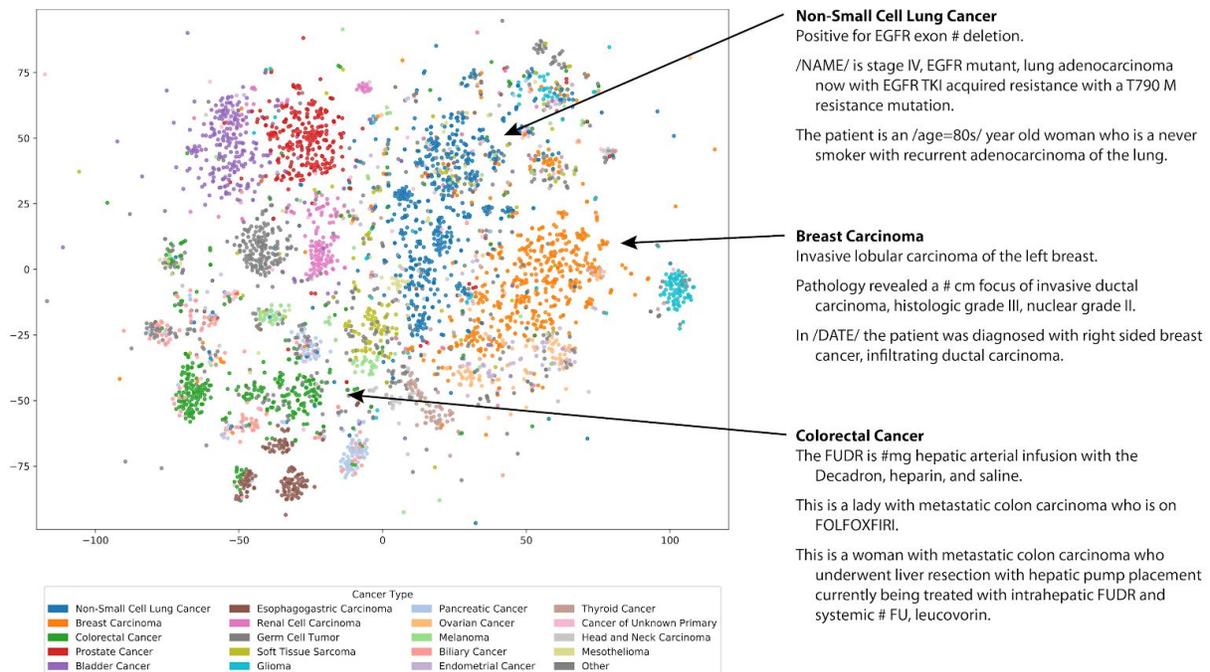

**Figure 2:** t-SNE embeddings[24] of patients represented by expressed word sentence cluster features. Each point is a patient, represented by the set of word sentence clusters detected within their text and are colored by their cancer type as determined by a pathologist (information not explicitly used). The median sentence of over-represented sentence clusters within each cancer type are shown on the right. De-identified tokens are shown with slashes (/).

## Somatic Association Study Results

The extracted phenotypes allow us to perform an association study with somatic mutations in tumors of the same patients in order to identify new statistical relationships between clinical features and somatic mutations. Somatic mutations in 340 genes were detected in tumors of each patient, as described in Zehir et al[19]. Most of them are very rare, so to increase the prevalence of these mutations, we collapse mutations to the gene level. We use the cancer type, number of documents and presence of Lynch Syndrome as covariates in a generalized linear model. We consider the three types of phenotypes for the association study:
    a) the CUI phenotypes represent patients as the set of CUI tokens detected in their EHR,
    b) the CUI Sentence Clusters (CSC) features represent patients as the set of CUI sentence clusters extracted from their EHR, and, analogously,
    c) the Word Sentence Cluster (WSC) features represent patients as the set of word sentence clusters extracted from their EHR.

That is, all sentences are clustered and each source sentence in the EHR is represented by its cluster membership. Associations are detected using a logistic regression model, accounting for confounding effects such as cancer type as linear covariates (see Methods).

Significant Association Analysis and Annotation Scheme

We find a total of 340 significant associations (FDR<5%): 202 associations with CUI token features, 69 associations with CUI sentence cluster features, and 69 associations with Word sentence cluster features. QQ and Volcano plots for the association studies are available in Supplemental Figures 3 and 4. The results of the association study are represented in **Figure 3** (full table available on request). Here genes are connected to phenotypes if there is a significant association between them. To demonstrate the power of sentence cluster phenotypes to capture multiple concepts, we attach unassociated CUI tokens to similar sentence clusters. Notions of similarity between CUI tokens and sentence clusters are discussed in the supplement. Details for this choice are found in the supplement, see Supplementary Figure 2. Also, genes are connected to nodes representing pathways that were found to play a role in many cancers[25]. These pathways are used for organizational purposes and do not confer any notions of gene-enrichment, as the genes selected for the study were already biased towards cancer-related genes.

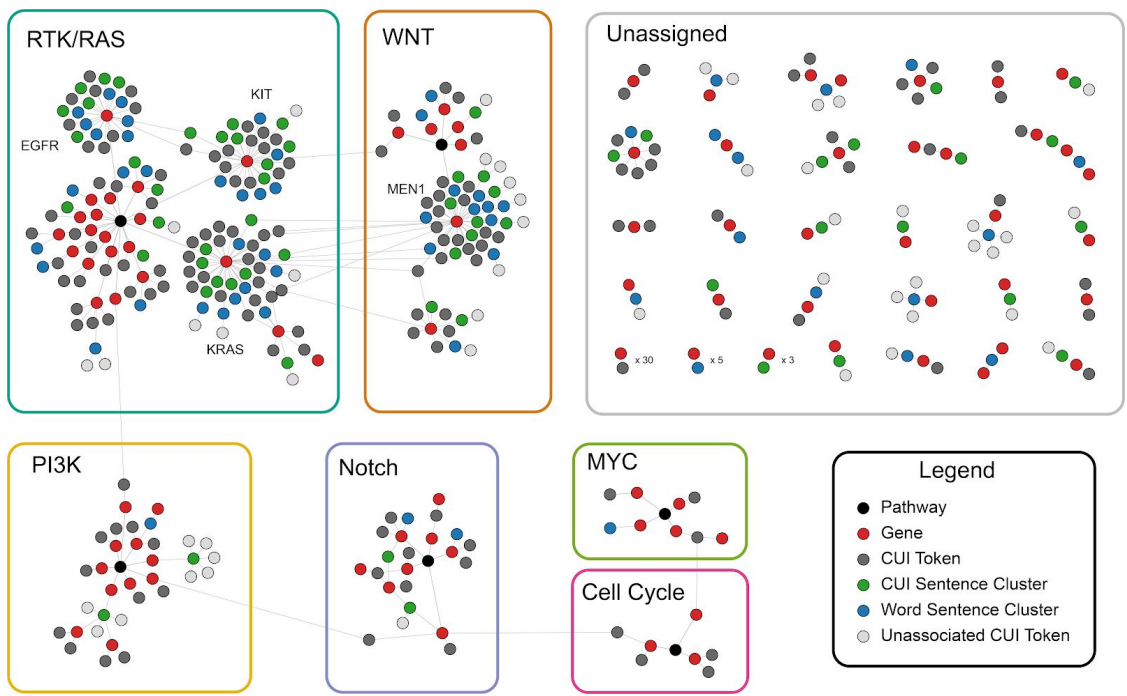

**Figure 3:** Graphical representation of the association study results. Each node represents a genotype, phenotype or pathway. Genotypes are connected to phenotypes if we identified a statistically significant association. Genes are connected to pathways they participate in. Bounding boxes organize the graph by the pathways. Unassociated CUI tokens are connected to sentence cluster phenotypes, if the jaccard similarity of the sets of patients expressing the features is above 20%.

## Annotations of Associations

Associations were annotated by an oncology expert into three categories according to their novelty. These categories are defined as *Known*, *Hypothesized*, and *Unknown*. The *Known* class includes associations for which there exists a well accepted explanation of the clinical or biological mechanism responsible for the association. This includes trivial associations, such as associating a gene to a phenotype describing the gene or its mutations, or associations with phenotypes representing diagnostics made or therapies administered by clinicians after detecting the mutation. The *Hypothesized* class contains associations for which there is not an accepted mechanism explaining the association, but we have proposed plausible and testable hypotheses (Table 2). The *Unknown* class contains associations which could not be rationalized. These include spurious or potentially false positives, or associations with phenotypes that are vague or otherwise difficult to interpret.

**Figure 4** shows the distribution of annotation labels per phenotype and pathway. Of the annotated associations, 52% of the associations made were already known, 12% were identified as potentially interesting and the remaining 36% were marked as unexplainable. While most associations were made with CUI tokens, the CSC associations had far fewer unexplainable mechanisms. This is likely due to clinical observations that necessitate multiple CUI codes. A single CUI token may capture part of this clinical observation, but when seen in isolation the interpretation of its mechanism becomes obfuscated. Relative to CSC features, WSC features have an increase in unexplainable associations, which is likely due to the increased difficulty of interpretation, since concepts can be spread over larger phrases, and clusters can become entangled. Additionally, with no mapping onto medical concepts, these clusters can capture non-medical concepts, for example administrative effects. However, this increase in unexplainable associations is only slight. Next, we group associations by the considered pathways[25] in which the genotypes participate. We offer this as a resource to biologists studying these pathways to help identify associations they may find interesting or relevant. Finally, we show the four most commonly associated genes and note that these are enriched for associations with known mechanisms, as these genes are well studied and already have therapies that target them.

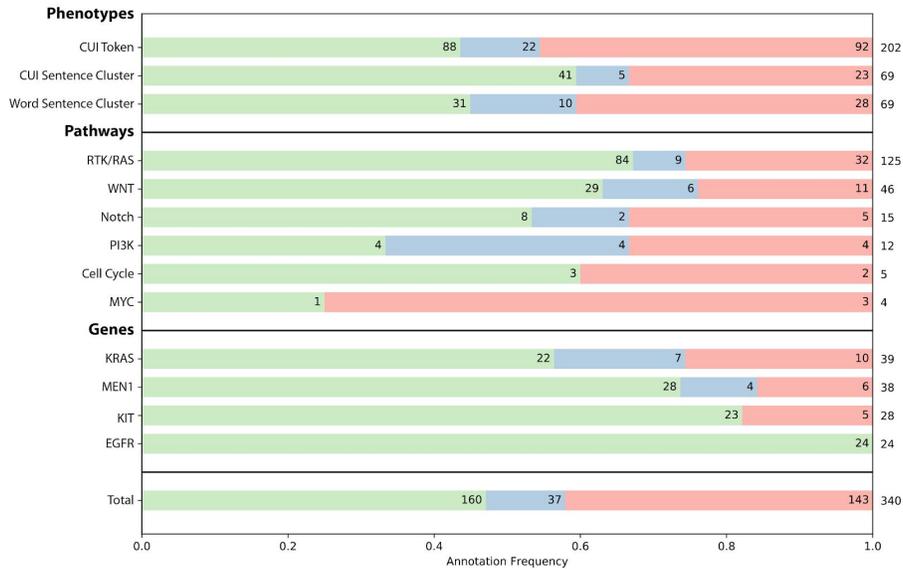

**Figure 4:** Distribution of association annotations (green: known associations; blue: hypothesized; red: unknown). Associations are grouped by their phenotype class, pathway and gene, for the four genes with the most associations. The number on the right hand side show the total number of associations for the group.

### Highlighted Associations

**Table 1** highlights examples of Known mutations. It is sorted first by phenotype class and then by significance. The mutations that occur in the highlighted CUI Token associations are used to inform treatment, and thus have strong and highly significant effects.

| Example Known Associations | | | | |
|---|---|---|---|---|
| **Genotype** | **Phenotype** | **Feature class** | **PV** | **β** |
| EGFR | [Erlotinib] | CUI | 1.53e-24 | 2.82 |
| EGFR | [Doxycycline] | CUI | 3.08e-07 | 1.25 |
| CDH1 | [Carcinoma, Lobular] | CUI | 1.30e-17 | 2.72 |
| EFGR | [Resistant] [Acquired] [Erlotinib] | CSC | 6.46e-07 | 3.54 |
| MEN1 | [Percent Positive Cells] [Radiolabeled Somatostatin Analog Study] | CSC | 9.28e-12 | 3.05 |
| ERBB2 | Neratinib protocol /TRIALNUM/. | WSC | 5.94e-10 | 3.14 |
| GNAS | Well differentiated mucinous adenocarcinoma of the appendix. | WSC | 2.52e-08 | 2.24 |
| MEN1 | She has had no diarrhea or flushing. | WSC | 8.37e-07 | 2.21 |

**Table 1:** Here we highlight a few significant associations tagged as Known. For these associations, there exists a well-known or agreed-upon explanation of the underlying biological or clinical mechanism. Sentence cluster phenotypes are represented by their median sentence. CUI tokens are indicated with square brackets. PV refers to the uncorrected p-value and β refers to the effect size of the association.

| **Example Hypothesized Associations** | | | | |
|---|---|---|---|---|
| Genotype | Phenotype | Feature class | PV | β |
| RICTOR | [Bilateral breast implants] | CUI | 7.90e-07 | 2.96 |
| Are (bilateral) breast implants associated with RICTOR (mTORC2) mutations in cancer?<br>Here we identify a testable hypothesis linking an intervention (breast implants) to a specific gene alteration. | | | | |
| CASP8 | [Peritoneal Neoplasms] | CUI | 4.19e-07 | 2.23 |
| Do CASP8 mutations augment the chance for peritoneal cancers or lead to more peritoneal metastasis of a non-peritoneal primary tumor? This would be a hypothesis which can be tested by looking at the data (were these true peritoneal primaries in these patients or metastatic tumors?).<br>Here, we identify a testable hypothesis linking a specific gene to clinical features of a tumor. | | | | |
| ZRSR2 | [Radiotherapy, Adjuvant] | CUI | 1.79e-08 | 2.57 |
| ZRSR2 is mutated in myelodysplastic syndromes. Radiotherapy increases the risk for myelodysplastic syndromes. Is this a mechanistic link?<br>Here, we identify a testable hypothesis for a mechanistic link between radiotherapy and myelodysplastic syndromes. | | | | |
| MAP2K4 | [Weight decreased] [Decrease in appetite] | CSC | 1.16e-06 | 2.84 |
| Could MAP2K4 mutations be linked to tumor cachexia?<br>Here, we identify a testable hypothesis associating MAP2K4 mutations to a specific cancer-related symptom (cachexia). | | | | |
| FANCA | [Hepatitis C, Chronic] | CSC | 1.19e-06 | 2.84 |
| FANCA mutations lead to anemia. Anemia may require erythrocyte transfusions. Transfusions carry an increased risk of hepatitis C transmission. Is this a link between FANCA mutations and hepatitis C?<br>Here, we identify a testable hypothesis which links FANCA mutations to chronic hepatitis C infection. | | | | |
| SPEN | Mohs surgery for basal cell carcinoma and squamous cell carcinoma | WSC | 4.25e-07 | 1.61 |
| Moh's surgery is a technique used for resection of skin cancer. Are SPEN mutations associated with skin cancer and thus with Moh's surgery?<br>Here, we identify a testable hypothesis linking the SPEN gene with the probability of a surgical intervention (Moh's surgery). | | | | |
| CTNNB1 | Child Pugh score A. | WSC | 1.44e-06 | 2.23 |
| CTNNB1 is a catenin, catenins are involved in liver cancer pathogenesis, Child Pugh Score rates liver function (amongst others in patients with liver cancer).<br>Here, we describe a testable hypothesis correlating the CTNNB1 gene with a clinical and lab index for liver function. | | | | |

**Table 2:** Here we highlight a few significant associations tagged as Hypothesized. In these cases, there does not exist a well known or agreed upon explanation of the underlying biological/clinical mechanism, however we provide a plausible and testable hypotheses for such a mechanism. Such results can be used by oncologists/cancer biologists to vet possible research directions. Sentence cluster phenotypes are represented by their median sentence. CUI tokens are indicated with square brackets and de-identified tokens are shown with slashes (/). PV refers to the uncorrected p-value and β refers to the effect size of the association.

**Table 2** highlights detected, potentially interesting novel associations and provides exemplary testable hypotheses to explain the underlying biological or clinical mechanism of the association.

## Discussion

In this work, we extract expressive patient representations from a large text corpus of unstructured clinical narratives. Our aim is to identify novel hypotheses about associations between a genotype and a phenotype. To achieve this, we derive a novel description for a phenotype based on unsupervised analysis sentences in EHRs: sentences, that represent a singular observation of a clinician, are clustered, and patients are represented as a set of the sentence clusters contained in their corresponding text.

We create two sets of sentence cluster features, defining sentences as bags-of-tokens over two vocabularies. The first vocabulary is an ontologic one, comprised of detected CUI tokens that represent medical concepts contained in the UMLS Ontology. These sentence clusters directly capture medical concepts detected in the text and showcase the ability of the sentence clusters to automatically detect meaningful combinations of medical concepts. The second vocabulary is comprised of natural language words. These features showcase the ability of the sentence clusters to capture useful information even when a resource such as an ontology is not available. We show that these sentence clusters detect complex medical observations, automatically and in an unsupervised manner, and contribute information that singular ontologic tokens could not capture.

We demonstrate the usefulness of our extracted features by using them as phenotypes in a genetic association study. Data of somatic mutations of cancer-related genes detected in tumors are matched to patients of the EHR corpus. Supplementing a patient phenotype representing patients as the set of ontologic concepts detected in their EHR with the two sentence cluster based phenotypes, we detected 340 significant associations, while correcting for confounding factors such as the cancer type of the patient. These associations were expert-annotated according to their novelty, where we show that the most significant associations describe well-known causes and effects of the associated mutation. We demonstrate that the meaning of the sentence cluster features were easily interpreted and contributed interesting information that the ontologic phenotype could not express. Additionally, we proposed plausible and testable hypotheses for the biological mechanisms of a subset of currently unexplainable associations. These associations could be used by researchers to guide experimental design, defining avenues for studying the effects of somatic mutations in advanced stage tumors, potentially informing therapies to target the causes or effects of these mutations. Additionally, these results can help inform clinical decision making and screening processes by identifying associated phenotypes describing risks and complications based on the genetics specific to the patient tumor.

Applications of these sentence cluster features are not limited to phenotypes in a genetic association study. The ability to query and study similar cases has been shown to be an effective strategy when a clinician is presented with a difficult decision[26]. In addition, the ability to find groups of patients expressing certain traits streamlines clinical trial selection[3,27]. Using sets of sentence clusters to represent patients and/or documents, clinicians and other healthcare staff would be able to query large EHR corpora in an easy and intuitive way. Sentences written in natural language can be provided as input to the query, which could then be parsed and assigned to sentence clusters, returning patients and/or documents that contain the detected features. Furthermore, the queries can be enhanced using the results of the association study. Matching genetic information to patients can increase the precision of the query results. Attaching associated genetic mutations to feature representations can aid a clinician interpretation and help suggest relevant unconsidered phenotypes.

As EHR adoption becomes more commonplace, larger text corpuses will be available, and our method will thus be able to find descriptive and specific features. Hierarchical clustering methods, such as the Louvain algorithm[28] used in this study, are able to provide granular and general phenotypes by producing phenotypes from each layer of the hierarchy. Additionally, access to larger EHR corpora could afford the use of deep learning methods. Word embeddings[29], or CUI embeddings[30], can be used to model token-token similarities whereas the Jaccard similarities used in this study assumes zero similarity between non-identical tokens. While CUI embeddings have been released as a resource, these were trained on a large EHR corpus from an insurance claims of general patients. As a result codes specific to cancer, which occur frequently in the EHR used for this study, are not well represented in the general EHR corpus, and thus have poor or missing representations, see Supplementary Figure 5. If a large cancer-specific EHR dataset would be available, the embeddings for these codes can be improved by using the pre-trained embeddings provided by Beam et. al.[30] and training them further using text from this larger cancer corpus. With such improved embeddings, one could also obtain embeddings for complete sentences[31,32]. These sentences could be clustered via their embedding representations, which offer more scalable clustering strategies, since in this setting the representations exist in a denser space of lower dimension. Such an approach may yield more informative and finer-grained features than the ones we obtain with our proposed model, but would require a significantly larger cancer text corpus.

Finally, as genetic sequencing becomes cheaper, sequencing cancer patients will become more commonplace and such genetic datasets will grow in size (see, for instance, the AACR Genie dataset). In this study, somatic mutations were collapsed to the gene level for statistical reasons, but larger cohorts would allow for more granular genotypes. This would help increase the power of the association study, whereas now the signal of mutational effects are potentially clouded by effects originating from other locations of the gene, when they are collapsed onto the gene level. Groupings based on the local neighborhoods of the mutation in the gene, or known/inferred mutational effect on the gene could be promising genotype options when considering larger cohorts.

## Methods

### Study design and setting

The study was designed as a retrospective cohort study for the development and analysis of techniques to extract features from clinical narratives and for association with somatic mutations. The study was performed at Memorial Sloan Kettering Cancer Center (MSKCC). The institutional review board of MSKCC provided a Waiver of Authorization (WOA; WA0426-13) for this study. Clinical notes were provided by the IT services group at MSKCC. Somatic mutations from the MSK Impact panel were provided by the Center for Molecular Oncology. We included all patients for which we had MSK Impact panel data and at least one clinical narrative available. Initial data analysis was performed on the HPC systems at MSKCC. Analyses were completed on a secured HPC system (Leonhard Med) provided by the IT Services at ETH Zürich, Switzerland.

### Text Processing & Normalization

We processed the raw EHR text with two separate parsers, which we refer to as the Word-parser and the CUI-parser. The Word-parser parses text to a cleaner natural language vocabulary. It maps tokens to lowercase, splits words backslashes and hyphen characters and removes special characters and punctuation. All numerical tokens are collapsed to the "#" character. Then the top 70 most frequent words over the corpus are removed as stop words. Words that occur less than 20 times are considered rare and are also removed. The CUI-parser parses text into the CUI code ontology, by using the MetaMap parser[33] to detect codes contained in the sentences. Each document in the corpus is processed with MetaMap[33], and a duplicate corpus is created where each sentence is replaced with its detected CUI codes. As in the Word-parser, the 70 most frequent codes and codes that appear less than 20 times are removed. Each sentence is then represented as a bag-of-tokens, where tokens are words under the Word-parser and CUI codes under the CUI-parser.

### Clustering Approach

In an effort to capture complex features, we cluster the sentences obtained in the previous step, creating two new sets of features which we refer to as a) Word Sentence Clusters (WSC): clusters derived from the Word-parsed sentences and b) CUI Sentence Clusters (CSC): clusters derived from the CUI-parsed sentences. Our clustering strategy consists of conducting a *k*-nearest neighbor (kNN) search to construct a kNN graph, which is then clustered using the Louvain method[28]. The kNN search identifies the *k* most similar sentences for each unique sentence, where similarity is measured using a weighted Jaccard scheme.

Given a vocabulary $\mathcal{V}$ and a set of sentences $\mathcal{S}$, sentences $\mathcal{A}, \mathcal{B} \in \mathcal{S}$ are represented as sets of tokens, $\mathcal{A}, \mathcal{B} \subset \mathcal{V}$. Their Jaccard similarity[34] is defined as:

$$\phi(\mathcal{A}, \mathcal{B}) = \frac{|\mathcal{A} \cap \mathcal{B}|}{|\mathcal{A} \cup \mathcal{B}|}$$

The weighted Jaccard scheme assumes a set of token weightings $\{w(t) : t \in \mathcal{V}\}$, and weighted set sizes are defined as $|\mathcal{A}| = \sum_{t \in \mathcal{A}} w(t)$. We weight tokens by the log of their inverse sentence frequency,

$$w(t) = log \frac{N}{n(t)}$$

where N is the number of unique sentences and n(t) is the number of sentences that contain token *t*. Other weightings were considered and are discussed in the supplement, refer to Supplementary Figure 1.

Scalable kNN Search

We employ an exact yet efficient kNN search over sentences. Our problem setting is concerned with high-dimensional (i.e., tens of thousands, depending on the size of the corpus) sparse binary feature vectors. Each query proceeds by computing the similarity of a sentence to all its neighbors, sorting these values, and finally returning the top *k* most similar. To scale to the dataset, we take advantage of the sparseness of the sentence matrix by only computing the similarity between sentences if their similarity is non-zero, i.e. they share at least one token. The removal of frequent words decreases the probability that two randomly selected sentences will have an non-empty intersection, thus restricting the number of similarities we need to compute and ultimately sort. Finally, the similarity computation for each sentence is independent and can be performed in parallel. Since the sentence matrix is stored efficiently, each process does not consume many resources.

Louvain Graph Clustering

A sentence kNN graph is constructed and clustered using the Louvain method[28]. Given the kNN of each sentence, we form a graph where each node represents a sentence, connected to each of its *k* neighbors. Edges are weighted by the similarity of the two sentences. The Louvain algorithm is a fast neighborhood detection algorithm that hierarchically clusters nodes in a graph and it has been shown to scale to graphs with upwards of 100 million nodes[28,35,36]. This algorithm is applied to the sentences from the CUI-parser and the Word-parser, creating the CUI Sentence Cluster (CSC) and Word Sentence Cluster (WSC) features. We set *k*=50 and use the first level of the hierarchy. Other choices for *k* were considered and are discussed in the supplement, see Supplementary Table 1.

**Association Model**

We use a logistic regression model to detect associations. To increase statistical power and reduce artifacts and unwanted associations, we consider patients with differing cancer types jointly, accounting for cancer type by including it as a covariate. This allows the model to represent the effect of the cancer type on the phenotype, thus reducing such spurious associations.

For cohort size $M$, a single phenotype $\vec{y} \in \{0,1\}^M$, a single genotype $\vec{x} \in \{0,1\}^M$ and $D$ covariates, each $\vec{z}_d \in \mathcal{R}^M$ and effect sizes $\alpha, \beta, \gamma_d \in \mathcal{R}$, we model

$$\vec{y} \sim \text{Bernoulli}(\sigma(l(\vec{x})))$$

where $\sigma$ denotes the sigmoid function:

$$\sigma(x) = \frac{e^x}{1-e^x}$$

And *l* is known as the log-odds and is defined as:

$$l = \alpha + \beta\vec{x} + \sum_{d=1}^{D} \gamma_d \vec{z_d}$$

We include the patient's cancer type, number of document and presence of Lynch Syndrome as covariates. Lynch Syndrome is an autosomal disorder that acts as a precursor to many cancer types, when left unaccounted it produced many undesired associations with it and its symptoms. We detect the presence of Lynch Syndrome in a patient by the occurrences of its CUI code. Cancer types are included if 50 or more patients exhibited them. The number of documents is included to explain away spurious correlations with administrative phenotypes that are more likely to appear in larger patient corpora.

Model parameters are learned via maximum likelihood. The value and p-value of **β** are used to describe the effect size and significance of the association. We define a minimum feature occurrence threshold, removing features below this value, as well as a rarity threshold, marking features below this value as rare. We set the minimum feature threshold to 10 patients and the rare feature threshold to 100 patients. All genotype-phenotype pairs were associated, omitting tests between a rare phenotype with a rare genotype. Since multiple tests for each feature were performed, p-values are corrected using the Benjamini-Hochberg method[37], which estimates the false discovery rate (FDR). We call an association significant, if its estimated FDR is less than 0.05. Additionally, we test the validity of the null hypothesis by performing associations where the genotype matrix is permuted. Parameter selection for the multiple testing correction is discussed in the supplement, see supplementary Table 3.

Acknowledgements:

This study was supported by the MSK Cancer Center Support Grant (P30 CA008748). Funding from this work was provided by the Sloan Kettering Institute (core funding to G.R.) and by ETH Zürich (core funding to G.R.). This work was partially supported by the European Union 7th Framework Programme through the Marie Curie Initial Training Network "Machine Learning for Personalized Medicine" MLPM2012 Grant No 316861. We gratefully acknowledge helpful discussions with Theofanis Karaletsos, Chris Sander, and Niki Schultz. Moreover, we would like to thank Iker Huerga, Chris Crosbie, Stuart Gardos and David Artz for supporting the work with data deliveries from the clinical data warehouse.

# Supplementary Material

## Clustering Parameters

There are two parameters to tune for the clustering pipeline: number of neighbors k and the form of similarity measure.

Three weighting schemes are considered, the ordinary Jaccard similarity *Basic*, inverse sentence frequency *ISF*, which places more emphasis on rare words, and *logISF* which eases this emphasis by applying a log transformation to the *ISF* weights. They are defined as:

$$w(t) = \begin{cases} 1, & \texttt{Basic} \\ \frac{N}{n(t)}, & \texttt{ISF} \\ log\frac{N}{n(t)}, & \texttt{logISF}, \end{cases}$$

Note that we would like the sentence cluster features to contribute information that cannot be expressed by CUI tokens alone, that is we aim to create features that capture concepts requiring multiple CUI tokens. To describe the number of CUI concepts a cluster captures, we use the number of tokens in the median sentence of CSC features as a proxy. The cluster median is sentence in the cluster with the smallest average distance to all other sentences in the cluster. This does not work for the WSC features, since they are based in natural language, where multiple words can be used to describe a single concept. **Supplementary Table 1** displays the number of CSC features whose median sentence contains more than one token across different weighting schemes and number of neighbors *k*. The *logISF* Jaccard weights and *k*=50 produced the largest number of novel features, and thus the downstream analysis is focused on these parameters.

| k | Basic | ISF | logISF |
|---|---|---|---|
| 10 | 2,639 | 1,368 | 2,399 |
| 50 | 2,489 | 2,030 | **3,163** |
| 100 | 1,622 | 1,868 | 1,974 |

**Supplementary Table 1:** Number of Novel CUI Sentence Clusters per clustering parameters. Here we show the number of CUI sentence clusters whose median sentence contain more than one CUI code. Ideally, sentence clusters capture combinations of CUI concepts, as this contributes information that individual CUI codes cannot supply. The number of tokens in the median CUI sentence cluster is used as a proxy for the number of concepts that cluster captures. CUI sentence clusters whose median sentence contains more than one CUI token are thus considered novel and we choose the clustering parameters, *logISF* Jaccard weights and k=50, which maximize the number of novel sentence clusters.

For an intuitive explanation of why this weighting scheme is preferred, consider the sentences

"The patient is currently taking Erlotinib" and "The patient is currently taking Neratinib." Erlotinib and Neratinib are names of drugs used to treat different conditions. The *Basic* Jaccard similarity would assign high similarity to these two sentences, however the *ISF* and *logISF* schemes assign lower similarity, since the drug names are less common, they will have larger weights compared to the other words in the sentence. As a result, the *Basic* weighting could create a cluster with the central concept of "taking medications", which is quite vague. *ISF* places a large emphasis on the rare tokens, i.e. the drug names, and could create clusters centered solely around these concepts, creating features that are already captured by the CUI tokens. *logISF*, on the other hand, downplays this emphasis, and seems to better capture the context of these central tokens. **Supplementary Figure 1** histograms of the number tokens in the median sentence of CSC features for *k*=50 across each Jaccard weighting scheme. Here we see that *Basic* creates fewer and thus more general clusters, while *ISF* creates many clusters, of which most are centered on a single concept.

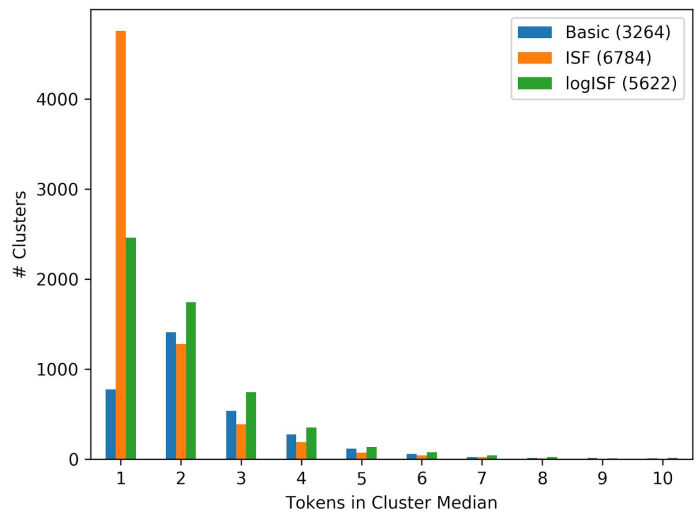

**Supplementary Figure 1:** Histogram of the number of CUI tokens in CUI sentence cluster median sentences for different Jaccard weightings. The number of CUI tokens in the cluster median is used as a proxy for the number of concepts the cluster captures. Ideally, sentence cluster medians contain more than one CUI token, since the information they capture can be expressed by simply using the CUI tokens. The *Basic* scheme produces general clusters, while the *ISF* scheme produces many clusters centered on a single CUI token. We opt for the *logISF* clusters as they created many specific phenotypes, but add information that is not captured by the CUI tokens alone.

| Cluster Type | # Unique Input Sentences | # Clusters | Median (Mean) Cluster Size | Median (Mean) Tokens in Cluster Center |
|---|---|---|---|---|
| **CUI Sentence Cluster** | 792,023 | 5,622 | 84 (122) | 2 (2.1) |
| **Word Sentence Cluster** | 1,009,062 | 4,757 | 90 (195) | 3 (4.0) |

**Supplementary Table 2**: Overview of the two sets of sentence clusters.

## Sentence Cluster Novelty

Next, we further develop the notion of novelty between features of different classes. Previously, we defined CSC features that had multiple tokens in their median sentence as novel, compared to CUI tokens. However, this does not generalize to WSC features, since multiple words are needed to explain some CUI tokens. Here we compare the source sentence sets for each feature to define similarities, $\phi$, across phenotype classes. Each sentence in the original corpus is mapped to a feature. For the CUI features, this mapping is many-to-many, since sentences can contain multiple CUI tokens, but for CSC and WSC, this mapping is many-to-one, since each sentence is assigned to at most one cluster. To measure the similarity between two features, we use the Jaccard similarity between their source sentence sets.

We define a threshold $\phi$ value, and say that two features are redundant if their similarity falls below this threshold. This will help communicate how much novelty each feature class contributes. We accomplish this by using $\phi$ to classify if a CSC feature captured multiple concepts, as we did previously. Supplementary Figure 2 shows the distribution of CSC features that capture a single CUI concept (their median sentence consists of a single token) and CSC features that capture multiple CUI concepts (their median sentence consists of multiple tokens) against the similarity of the closest CUI feature. From this, we define features for which $\phi > 0.2$ as redundant.

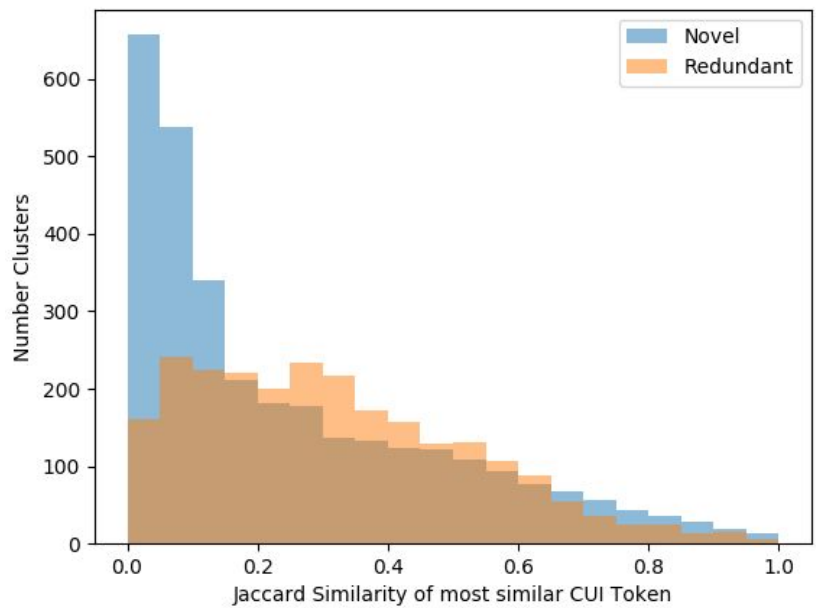

**Supplementary Figure 2**: Histogram of the similarity between CUI Sentence Clusters and the most similar CUI token. Similarity is measured by the Jaccard similarity between the sets of patients expressing each feature. With this strategy, we are able to compare two features from different phenotype classes. CUI Sentence Clusters are considered Novel if the median sentence of the cluster contains more than one CUI token. If the median sentence contains only one token, then it is considered redundant to that CUI token. Using this split we conclude that $\phi = 0.20$ is a good threshold value for measuring redundancy between feature classes.

## Association Parameter Selection

While associations are performed independently pairwise between all genotypes and phenotypes, multiple testing correction methods operate on the entire set of experiments. Thus, the behavior of the corrected p-values will depend on the entire feature set, and choices regarding the parameters determining this feature set must be made to keep the experiment well-behaved. We define and tune two thresholding parameters that select testable genotype-phenotype pairs. The first threshold, *Min*, removes omits features that occur in less than *Min* patients. The second threshold, *Rare*, marks features that occur in less than *Rare* patients as "rare," and tests between a rare genotype and a rare phenotype are omitted. The third parameter, *QV*, determines the threshold at which the corrected p-values are considered significant. These parameters are tuned to maximize the total number of association, while keeping the experiment well behaved. To confirm an experiment is well behave, we estimate the true false positive rate by performing associations on a permuted phenotype patient set, applying the same permutation to all features. In well behaved experiments, the number of associations reported by the permuted data should reflect the false discovery rate. **Supplementary Table 3** summarizes the number of true and permuted associations at each parameter setting. Ultimately, we chose *Min*=10, *Rare*=100, *FDR*=0.05 as this produced well-behaved experiments with a large amount of true associations.

| Min | Rare | FDR | CUI | | CSC | | WSC | | Total | |
|---|---|---|---|---|---|---|---|---|---|---|
| | | | True | Pmt | True | Pmt | True | Pmt | True | Pmt |
| **10** | 10 | 0.01 | 100 | 13 | 39 | 6 | 24 | 7 | 163 | 26 |
| | | 0.05 | 627 | 1110 | 127 | 174 | 101 | 149 | 855 | 1433 |
| | | 0.1 | 1922 | 4101 | 365 | 610 | 263 | 553 | 2550 | 5264 |
| | 50 | 0.01 | 81 | 3 | 40 | 5 | 26 | 0 | 147 | 8 |
| | | 0.05 | 318 | 196 | 105 | 127 | 87 | 97 | 510 | 420 |
| | | 0.1 | 722 | 1121 | 294 | 392 | 206 | 272 | 1222 | 1785 |
| | **100** | 0.01 | 82 | 0 | 30 | 3 | 29 | 0 | 141 | 3 |
| | | **0.05** | **202** | **12** | **69** | **14** | **69** | **2** | **340** | **28** |
| | | 0.1 | 360 | 155 | 164 | 48 | 156 | 59 | 680 | 262 |
| | 200 | 0.01 | 33 | 1 | 14 | 0 | 13 | 0 | 60 | 1 |
| | | 0.05 | 71 | 1 | 22 | 0 | 24 | 4 | 117 | 5 |
| | | 0.1 | 109 | 2 | 41 | 0 | 48 | 4 | 198 | 6 |

| | | | | | | | | | | | |
|---|---|---|---|---|---|---|---|---|---|---|---|
| 25 | 25 | 0.01 | 78 | 3 | 34 | 4 | 24 | 4 | 136 | 11 |
| | | 0.05 | 220 | 112 | 88 | 85 | 71 | 66 | 379 | 263 |
| | | 0.1 | 459 | 443 | 217 | 231 | 157 | 169 | 833 | 843 |
| | 50 | 0.01 | 75 | 2 | 32 | 3 | 24 | 1 | 131 | 6 |
| | | 0.05 | 220 | 112 | 88 | 85 | 71 | 66 | 379 | 263 |
| | | 0.1 | 375 | 254 | 198 | 155 | 144 | 156 | 717 | 565 |
| | 100 | 0.01 | 70 | 1 | 27 | 2 | 28 | 0 | 125 | 3 |
| | | 0.05 | 162 | 5 | 62 | 10 | 61 | 14 | 285 | 29 |
| | | 0.1 | 272 | 44 | 117 | 33 | 135 | 38 | 524 | 115 |
| | 200 | 0.01 | 36 | 1 | 16 | 0 | 14 | 0 | 66 | 1 |
| | | 0.05 | 69 | 2 | 25 | 0 | 25 | 3 | 119 | 5 |
| | | 0.1 | 101 | 2 | 46 | 0 | 59 | 9 | 206 | 11 |
| 50 | 50 | 0.01 | 49 | 2 | 14 | 2 | 12 | 2 | 75 | 6 |
| | | 0.05 | 114 | 11 | 36 | 11 | 30 | 14 | 180 | 36 |
| | | 0.1 | 189 | 44 | 75 | 55 | 72 | 61 | 336 | 160 |
| | 100 | 0.01 | 48 | 1 | 15 | 2 | 13 | 0 | 76 | 3 |
| | | 0.05 | 105 | 2 | 27 | 7 | 36 | 8 | 168 | 17 |
| | | 0.1 | 171 | 26 | 57 | 19 | 66 | 25 | 294 | 70 |
| | 200 | 0.01 | 31 | 1 | 12 | 0 | 8 | 0 | 51 | 1 |
| | | 0.05 | 55 | 1 | 14 | 0 | 17 | 2 | 86 | 3 |
| | | 0.1 | 67 | 1 | 19 | 0 | 27 | 8 | 113 | 9 |

**Supplementary Table 3**: Grid search over parameters used to determine significant associations via multiple testing correction. Features are removed if the total number of patients expressing them is below *Min*. Features are considered rare if the total number of patients expressing them is below *Rare*. All tests between a rare genotype and a rare phenotype are removed. Associations for which the false discovery rate produced by the Benjamini-Hochberg method[37] are less than *FDR* are considered significant. Ultimately, associations from *Min*=10 *Rare*=100 and *FDR*=0.05 were selected as they created a large number of significant associations, with well-behaved permuted associations.

## Association Quality Control

**Supplementary Figure 3** shows QQ plots for each experiment. The QQ plot is a scatter plot of observed p-values against expected p-value, on a negative log scaling. The true associations are shown in dark blue, while permuted associations are shown in light blue. Under the null hypothesis, p-values should follow a uniform distribution, visualized with the solid line. In a well behaved study, the permuted associations would follow the solid line, since all phenotype-genotype pairs under the permutation are expected to follow the null. Here we see slight inflation, which is reflected by the slightly increased false positive rate in the number of permuted associations. The inflation in the true could be explained by unaccounted for confounders, and necessitates expert input to help differentiate the true positives from the false positives.

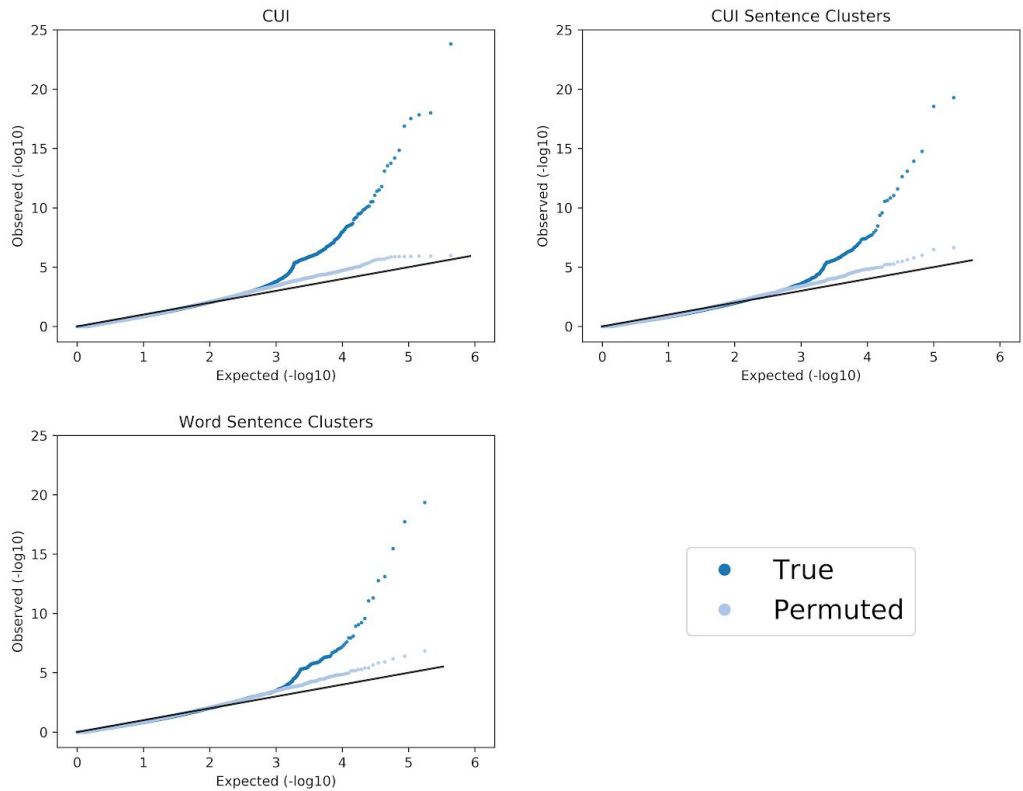

**Supplementary Figure 3**: QQ plots for each of the three association studies. P-values from the true associations are shown in dark blue, p-values from the permuted associations are shown in light blue. Ideally, uninflated permuted associations follow the diagonal, indicated by the black line.

**Supplementary Figure 4** reports volcano plots of each experiment. Each element in a volcano plot represents the effect size (β) and significance (p-value, negative log scaled) of an association. Significant associations are colored according to their annotation. *Known* associations are shown to be the strongest and most significant associations.

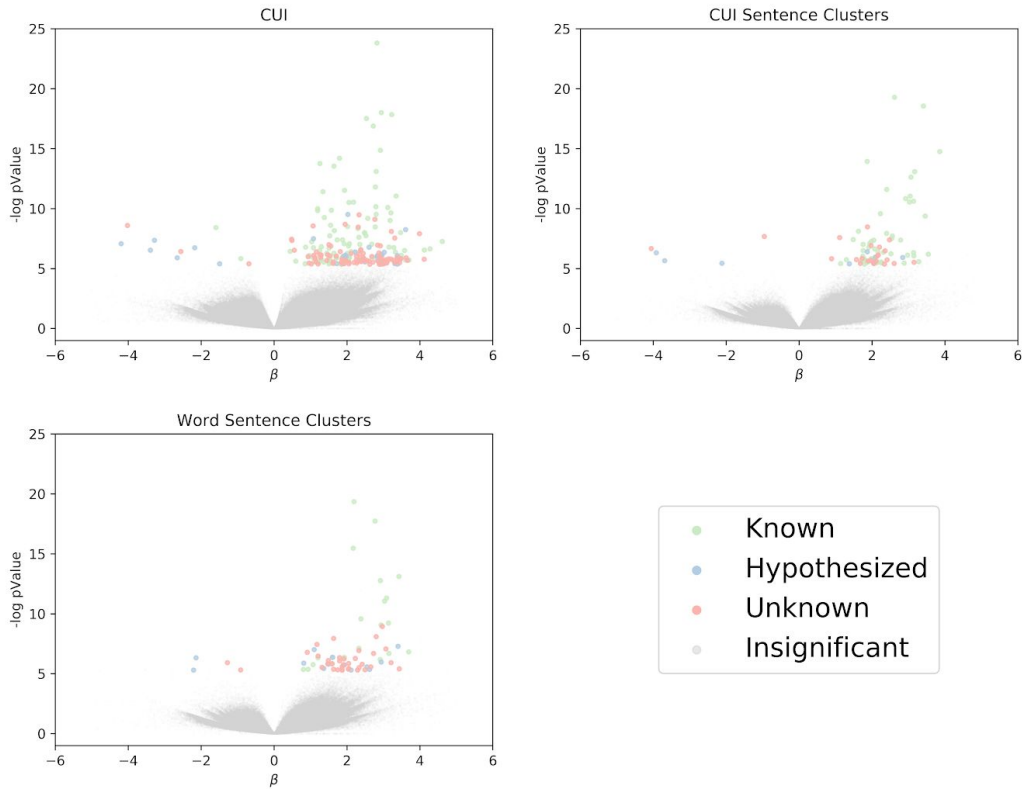

**Supplementary Figure 4:** Volcano plots for each of the three association studies. P-value of each association (negative $\log_{10}$ transformed) is scattered against the effect size of the association (β). From this we observe that our associations annotated as known are among the strongest and most significant associations.

## CUI Embeddings do not contain enough cancer-specific codes

We considered the use of embedded representations of CUI tokens, such as those produced by Beam et al[30]. However, these embeddings were trained on a dataset of general medical topics. While the size of this dataset is massive, our dataset is concerned specifically with cancer patients, and embeddings for many CUI tokens were missing. **Supplementary Figure 5** shows a histogram of present and missing embeddings against the rarity (i.e. the number of patients exhibiting the CUI token). We observe that many codes are missing, and independent of their frequency in the dataset.

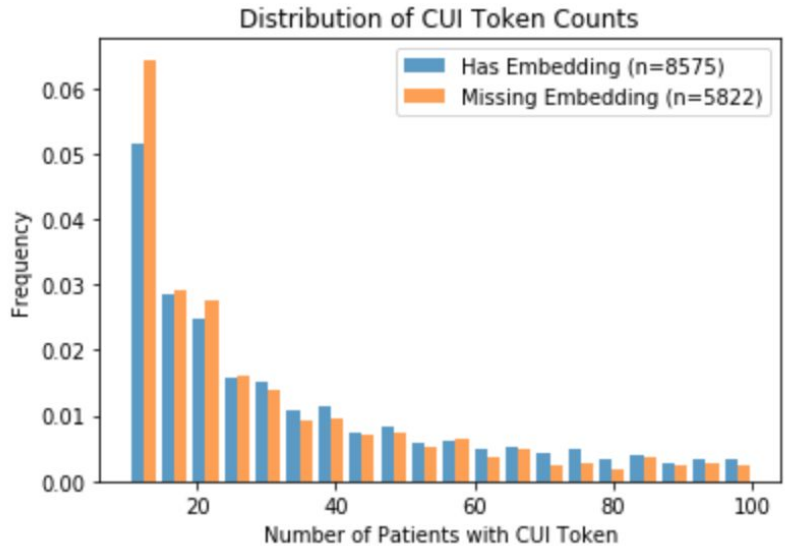

**Supplementary Figure 5**: Histograms of missing and present CUI embeddings as provided by Beam et al.[30] These embeddings were trained from a massive corpus of general medical records. We observe that CUI tokens are missing from our corpus independently from their rarity, suggesting that these codes are cancer specific. Use of these embeddings could introduce more powerful clustering strategies, however a larger cancer-specific text corpus is needed to learn proper embeddings.

# References

1. Denny, J. C., Bastarache, L. & Roden, D. M. Phenome-wide association studies as a tool to advance precision medicine. *Annu. Rev. Genomics Hum. Genet.* **17**, 353–373 (2016).
2. Kohane, I. S. Using electronic health records to drive discovery in disease genomics. *Nat. Rev. Genet.* **12**, 417 (2011).
3. Simmons, M., Singhal, A. & Lu, Z. Text Mining for Precision Medicine: Bringing Structure to EHRs and Biomedical Literature to Understand Genes and Health. in *Translational Biomedical Informatics: A Precision Medicine Perspective* (eds. Shen, B., Tang, H. & Jiang, X.) 139–166 (Springer Singapore, 2016).
4. Zhou, L., Suominen, H. & Gedeon, T. Adapting State-of-the-Art Deep Language Models to Clinical Information Extraction Systems: Potentials, Challenges, and Solutions. *JMIR Med Inform* **7**, e11499 (2019).
5. Wang, Y. *et al.* Natural language processing of radiology reports for identification of skeletal site-specific fractures. *BMC Med. Inform. Decis. Mak.* **19**, 73 (2019).
6. Yazdani, A., Safdari, R., Golkar, A. & R Niakan Kalhori, S. Words prediction based on N-gram model for free-text entry in electronic health records. *Health Inf Sci Syst* **7**, 6 (2019).
7. Savova, G. K. *et al.* DeepPhe: A Natural Language Processing System for Extracting Cancer Phenotypes from Clinical Records. *Cancer Res.* **77**, e115–e118 (2017).
8. Bodenreider, O. The Unified Medical Language System UMLS: integrating biomedical terminology. *Nucleic Acids Res.* **32**, D267–70 (2004).
9. Aronson, A. R. Effective mapping of biomedical text to the UMLS Metathesaurus: the

MetaMap program. *Proc. AMIA Symp.* 17–21 (2001).
10. Manolio, T. A. Genomewide association studies and assessment of the risk of disease. *N. Engl. J. Med.* **363**, 166–176 (2010).
11. Calabrese, C. *et al.* Genomic basis for RNA alterations revealed by whole-genome analyses of 27 cancer types. *bioRxiv* (2018).
12. Kahles, A. *et al.* Comprehensive Analysis of Alternative Splicing Across Tumors from 8,705 Patients. *Cancer Cell* **34**, 211–224.e6 (2018).
13. Bush, W. S., Oetjens, M. T. & Crawford, D. C. Unravelling the human genome--phenome relationship using phenome-wide association studies. *Nat. Rev. Genet.* **17**, 129 (2016).
14. Jensen, P. B., Jensen, L. J. & Brunak, S. Mining electronic health records: towards better research applications and clinical care. *Nat. Rev. Genet.* **13**, 395 (2012).
15. Sud, A., Kinnersley, B. & Houlston, R. S. Genome-wide association studies of cancer: current insights and future perspectives. *Nat. Rev. Cancer* **17**, 692–704 (2017).
16. Easton, D. F. *et al.* Genome-wide association study identifies novel breast cancer susceptibility loci. *Nature* **447**, 1087–1093 (2007).
17. Wang, Y. *et al.* Common 5p15.33 and 6p21.33 variants influence lung cancer risk. *Nat. Genet.* **40**, 1407–1409 (2008).
18. Wheeler, D. A. *et al.* The complete genome of an individual by massively parallel DNA sequencing. *Nature* **452**, 872–876 (2008).
19. Zehir, A. *et al.* Mutational landscape of metastatic cancer revealed from prospective clinical sequencing of 10,000 patients. *Nat. Med.* **23**, 703 (2017).
20. Denny, J. C. *et al.* Variants near FOXE1 are associated with hypothyroidism and other thyroid conditions: using electronic medical records for genome-and phenome-wide studies. *Am. J. Hum. Genet.* **89**, 529–542 (2011).
21. Hebbring, S. J. *et al.* Application of clinical text data for phenome-wide association studies (PheWASs). *Bioinformatics* **31**, 1981–1987 (2015).
22. Denny, J. C. *et al.* Identification of genomic predictors of atrioventricular conduction: using electronic medical records as a tool for genome science. *Circulation* **122**, 2016–2021 (2010).
23. Chan, K. R. *et al.* An Empirical Analysis of Topic Modeling for Mining Cancer Clinical Notes. in *2013 IEEE 13th International Conference on Data Mining Workshops* 56–63 (2013).
24. Maaten, L. van der & Hinton, G. Visualizing Data using t-SNE. *J. Mach. Learn. Res.* **9**, 2579–2605 (2008).
25. Sanchez-Vega, F. *et al.* Oncogenic Signaling Pathways in The Cancer Genome Atlas. *Cell* **173**, 321–337 (2018).
26. Lowe, H. J., Ferris, T. A., Hernandez, P. M. & Weber, S. C. STRIDE--An integrated standards-based translational research informatics platform. *AMIA Annu. Symp. Proc.* **2009**, 391–395 (2009).
27. Ni, Y. *et al.* Increasing the efficiency of trial-patient matching: automated clinical trial eligibility pre-screening for pediatric oncology patients. *BMC Med. Inform. Decis. Mak.* **15**, 28 (2015).
28. Blondel, V. D., Guillaume, J.-L., Lambiotte, R. & Lefebvre, E. Fast unfolding of communities in large networks. *J. Stat. Mech: Theory Exp.* **2008**, P10008 (2008).

29. Mikolov, T., Sutskever, I., Chen, K., Corrado, G. S. & Dean, J. Distributed Representations of Words and Phrases and their Compositionality. in *Advances in Neural Information Processing Systems 26* (eds. Burges, C. J. C., Bottou, L., Welling, M., Ghahramani, Z. & Weinberger, K. Q.) 3111–3119 (Curran Associates, Inc., 2013).
30. Beam, A. L. *et al.* Clinical concept embeddings learned from massive sources of medical data. *arXiv preprint arXiv:1804. 01486* (2018).
31. Pagliardini, M., Gupta, P. & Jaggi, M. Unsupervised Learning of Sentence Embeddings using Compositional n-Gram Features. *arXiv [cs.CL]* (2017).
32. Wieting, J., Bansal, M., Gimpel, K. & Livescu, K. Towards Universal Paraphrastic Sentence Embeddings. *arXiv [cs.CL]* (2015).
33. Aronson, A. R. Effective mapping of biomedical text to the UMLS Metathesaurus: the MetaMap program. in *Proceedings of the AMIA Symposium* 17 (American Medical Informatics Association, 2001).
34. Jaccard, P. New research on floral distribution. *Bull. Soc. Vaud. sci. nat.* **44**, (1908).
35. Haynes, J. & Perisic, I. Mapping Search Relevance to Social Networks. in *Proceedings of the 3rd Workshop on Social Network Mining and Analysis* 2:1–2:7 (ACM, 2009).
36. Pujol, J. M., Erramilli, V. & Rodriguez, P. Divide and Conquer: Partitioning Online Social Networks. *CoRR* **abs/0905.4918**, (2009).
37. Benjamini, Y. & Hochberg, Y. Controlling the false discovery rate: a practical and powerful approach to multiple testing. *J. R. Stat. Soc. Series B Stat. Methodol.* 289–300 (1995).